# Traffic Graph Convolutional Recurrent Neural Network: A Deep Learning Framework for Network-Scale Traffic Learning and Forecasting

Zhiyong Cui, Kristian Henrickson, Ruimin Ke, Ziyuan Pu, and Yinhai Wang*

*Abstract*— Traffic forecasting is a particularly challenging application of spatiotemporal forecasting, due to the time-varying traffic patterns and the complicated spatial dependencies on road networks. To address this challenge, we learn the traffic network as a graph and propose a novel deep learning framework, Traffic Graph Convolutional Long Short-Term Memory Neural Network (TGC-LSTM), to learn the interactions between roadways in the traffic network and forecast the network-wide traffic state. We define the traffic graph convolution based on the physical network topology. The relationship between the proposed traffic graph convolution and the spectral graph convolution is also discussed. An L1-norm on graph convolution weights and an L2-norm on graph convolution features are added to the model's loss function to enhance the interpretability of the proposed model. Experimental results show that the proposed model outperforms baseline methods on two real-world traffic state datasets. The visualization of the graph convolution weights indicates that the proposed framework can recognize the most influential road segments in real-world traffic networks.

## I. Introduction

TRAFFIC forecasting is one of the most challenging components of Intelligent Transportation Systems (ITS). The goal of traffic forecasting is to predict future traffic states in the traffic network given a sequence of historical traffic states and the physical roadway network. Since the volume and variety of traffic data has been increasing in recent years, data-driven traffic forecasting methods have shown considerable promise in their ability to outperform conventional and simulation-based methods [1].

Previous work [2][3][4][5] on this topic roughly categorizes existing models into two categories: classical statistical methods and machine learning models. Most of the studies focusing on traffic forecasting using statistical methods were developed when traffic systems were less complex, and the sizes of traffic datasets were relatively small. However, statistical models' capability of handling high dimensional time series data is quite limited. With the more recent rapid development in computational power, as well as growth in traffic data volume, much of the more recent work on this topic focuses on machine learning methods for traffic forecasting.

Machine learning methods with the capability of capturing complex non-linear relationships, like support vector regression (SVR) [6], tend to outperform the statistical methods, such as autoregressive integrated moving average (ARIMA) [7] and its variants, with respect to handling complex traffic forecasting problems [8]. However, the full potential of artificial intelligence approaches to traffic forecasting was not exploited until the rise of deep neural network (NN) models (also referred to as deep learning models). Following early works [2], [9] applying NNs to the traffic prediction problem, many NN-based methods have been adopted for traffic forecasting.

Deep learning models for traffic forecasting, such as deep belief networks (DBN) [10] and stacked auto-encoders [11], can effectively learn high dimensional features and achieve good forecasting performance. Recurrent neural network (RNN) and its variants, including long short-term memory (LSTM) [12] and gated recurrent unit (GRU) [13] networks, have also shown great potential for solving traffic forecasting problems [8], [14], [15], [16]. Although RNN-based methods can learn the spatial dependencies, they tend to be over-complex and inevitably capture a certain amount of noise and spurious relationships which likely do not represent the true causal structure in a physical traffic network. Moreover, interpreting the network parameters in terms of real-world spatial dependencies is most often impossible. To address this, other works [5], [17], [18] attempt to model spatial dependencies with convolutional neural network (CNN). However, conventional CNNs are most appropriate for spatial relationships in the Euclidean space as represented by two-dimensional (2D) matrices or images. Thus, spatial features learned in CNN are not optimal for representing the traffic network structure [19][20].

Recently, substantial research has focused on extending the convolution operator to more general, graph-structured data, which can be applied to capture the spatial relationships present in a traffic network. There are two primary ways to conduct graph convolution. The first class of methods [21], [22], [23],

Z. Cui, R. Ke, Z. Pu, and Y. Wang are with the Department of Civil and Environmental Engineering, University of Washington, Seattle WA 98195, USA (e-mail: zhiyongc@uw.edu; ker27@uw.edu; ziyuanpu@uw.edu; yinhai@uw.edu).

K. Henrickson is with the INRIX, Inc., Kirkland WA 98033, USA (e-mail: Kristian.Henrickson@inrix.com)

[24] makes use of spectral graph theory, by designing spectral filter/convolutions based on the graph Laplacian matrix. Spectral-based graph convolution has been adopted and combined with RNN [20] and CNN [1] to forecast traffic states. These models successfully apply convolution to graph-structured data, but they do not fully capture the unique properties of graphs [25], like traffic networks. These models [23], [26] usually adopt multiple graph convolution layers, and thus, their learned spatial dependencies are hard to interpret. The other form of graph convolution proposed in several newly-published studies is conducted on graph data dynamically, for example, the dynamic edge-conditioned filters in graph convolution [27], the high-order adaptive graph convolutional network [25][28]. Still, these methods are not capable of fully accommodating the physical specialties of traffic networks.

One of the deficiencies of the previous graph convolution-based models is that the receptive field of the convolution operators is not confined in the graph according to the real structure of the traffic network. The traffic states of two locations far apart from each other in the traffic network should not be influenced by each other in a short time period. Though the spectral graph convolution models [20],[23] can capture features from K-localized neighbors of a vertex in the graph, how to choose the value of K and whether the localized neighbors truly affect the vertex are still questions to be answered. Thus, we propose a free-flow reachable matrix based on the free-flow speed of the real traffic and apply it on the graph convolution operator to learn features from the truly influential neighborhood in the traffic network.

In this study, we learn the traffic network as a graph and conduct convolution on the traffic network-based graph. To learn localized features and incorporate roadway physical characteristics, we proposed a traffic graph convolution operator. Base on this operator, we propose a traffic graph convolutional LSTM (TGC-LSTM) to model the dynamics of the traffic flow and capture the spatial dependencies. Evaluation results show that the proposed TGC-LSTM outperforms multiple state-of-the-art traffic forecasting baselines. More importantly, the proposed model turns out to be capable of identifying the most influential roadway segments in the real-world traffic networks. The main contributions of our work include:
1. A traffic graph convolution operator is proposed to accommodate physical specialties of traffic networks and extract comprehensive features.
2. A traffic graph convolutional LSTM neural network is proposed to learn the complex spatial and dynamic temporal dependencies presented in traffic data.
3. To make learned localized graph convolution features more consistent and interpretable, we proposed two regularization terms, including an L1-norm on traffic the graph convolution weights and an L2-norm on the traffic graph convolution features, that can be optionally added to the model's loss function.
4. The real-world traffic speed data, including the graph structure of the traffic network, used in this study is published via a publicly available website[1] to facilitate further research on this problem.

II. LITERATURE REVIEW

A. Deep Learning based Traffic Forecasting

Deep learning models have shown their superior capabilities of capturing nonlinear spatiotemporal effects for traffic forecasting [29]. Ever since the precursory study [30] using the feed-forward NN for vehicle travel time estimation was proposed, many other NN-based models, including fuzzy NN [31], recurrent NN [9], convolution NN [5][18], deep belief networks [10][32], auto-encoders [11][33], generative adversarial networks [34][35], and combinations of these models have been applied to forecast traffic states. With the capability of capturing temporal dependencies, the recurrent NN or its variants, like LSTM [12] and GRU [13], was widely adopted as a component of a traffic forecasting model to forecast traffic speed [8], travel time [36], and traffic flow [37].

Further, in most recent years, various novel deep learning-based traffic forecasting models have been proposed through adjusting classical neural network model, combining existing methods, and incorporating auxiliary data. Multiple novel LSTM based models, such as bidirectional LSTM [14], deep LSTM [15], shared hidden LSTM [38], and nested LSTM [39], have been designed via reorganizing and combing single LSTM models and applied to capture comprehensive temporal dependencies for traffic prediction. In addition, sequence-to-sequence (seq2seq) architecture based models [20],[33] have also been used for traffic state sequence forecasting. To deal with different types of features, multi-stream deep learning models [15][40][41][42] have also been well studied and tested for traffic forecasting problems. To improve the prediction performance, multiple deep learning based models also incorporate various traffic-related auxiliary data, including roadway geographical attribute data [33], accident data [15], and weather data [43].

To capture spatial relationships present in traffic networks, many forecasting models [5], [44] incorporating CNNs to extract spatial features from 2D spatial-temporal traffic data. Due to the traffic structure is hard to be depicted by 2D spatial-temporal data, studies [18] tried to convert traffic network structures to images and use CNNs to learn spatial features. However, these converted images have a certain amount of noise, inevitably resulting in spurious spatial relationships captured by CNNs. Recent studies [42][45][46] also attempted to convert traffic state data into three-dimensional (3D) matrices and use the 3D convolutional network to extract more effective features. However, conventional CNN based methods still cannot inherently deal with the topological structure and the physical attributes of the traffic network. To solve this problem, studies [1], [20] attempted to learn the traffic network as a graph and adopt the graph-based convolution operator to extract features from the graph-structured traffic network.

---
[1] https://github.com/zhiyongc/Seattle-Loop-Data

## B. Graph Convolution Networks

Traffic networks have already been analyzed as graphs for dynamic shortest path routing [47], traffic congestion analysis [48], and dynamic traffic assignment [49]. In the last couple of years, many studies attempt to generalize neural networks to work on arbitrarily structured graphs by designing graph convolutional networks. Generally, the graph convolutional networks utilize the adjacency matrix or the Laplacian matrix to depict the structure of a graph. The Laplacian matrix based graph convolution [22], [26] are designed based on the spectral graph theory [50]. As an extension, a localized spectral graph convolution [23] is also proposed to reduce the learning complexity. The adjacency matrix based graph convolution neural networks [24], [25] incorporate the adjacency matrix and their network structures are more flexible. The traffic network can be considered as a graph consisting of nodes and edges, and thus, several graph convolution neural network based models, including the spectral graph convolution [1] and the diffusion graph convolution [21], are proposed to fulfill network-wide traffic forecasting. Several studies [51][52] also incorporated multi-scale graph convolution operations into their proposed models to learn traffic features. Although these existing methods can extract spatial features from neighborhoods in the traffic network, the physical specialties of roadways, like length, speed limits, and the number of lanes, are normally neglected.

## III. METHODOLOGY

### A. Notions

#### 1) Traffic Network based Graph

Normally, a graph consists of nodes (vertices) and edges. The graph representing a traffic network is distinct from social network graphs, document citation graphs, or molecule graphs, in several respects: 1) there are no isolated nodes/edges in traffic network based graphs and the traffic network structure seldom changes; 2) the traffic status of each road in a traffic network varies over time; and 3) the roads in a traffic network have meaningful physical characteristics, such as the length, type, speed limit, and lane numbers of a road. Further, traffic state data is collected by different types of sensors such that some types of data detect location-based traffic states, but others may measure road segment based averaged traffic states. Due to traffic states vary over time, it is better to let the graph nodes possess the varying traffic states and keep the graph structure fixed. Thus, to ensure the consistency of the definition in a graph, we use **nodes** to represent the traffic sensing locations, which can be sensor stations or road segments. Then, the **edges** in a graph represent the intersections or road segments connecting those traffic sensing locations.

The traffic network and the relationship between traffic locations can be represented by an undirected graph $\mathcal{G}$ where $\mathcal{G} = (\mathcal{V}, \mathcal{E})$ with $N$ nodes $v_i \in \mathcal{V}$ and edges $(v_i, v_j) \in \mathcal{E}$. Even though some roads are directed in the reality, due to the impact of traffic congestions occurring on these roads will be bi-directionally propagated to upstream and downstream roads [14], we take the bidirectional impact into account and thus let $\mathcal{G}$ be an undirected graph.

#### 2) Adjacency Matrix and Neighborhood Matrix

The connectedness of nodes in $\mathcal{G}$ is represented by an adjacency matrix $A \in \mathbb{R}^{N \times N}$, in which each element $A_{i,j} = 1$ if there is an edge connecting node $i$ and node $j$ and $A_{i,j} = 0$ otherwise ($A_{i,i} = 0$). Based on the adjacency matrix, the degree matrix of $\mathcal{G}$, which measures the number of edges attached to each vertex, can be defined as $D \in \mathbb{R}^{N \times N}$ in which $D_{ii} = \sum_j A_{ij}$. $D$ is a diagonal matrix and all non-diagonal elements are zeros.

Based on the adjacency matrix, an edge counting function $d(v_i, v_j)$ can be defined as counting the minimum number of edges traversed from node $i$ to node $j$. Then, the set of $k$-hop ($k$-th order) neighborhood of each node $i$, including node $i$ itself, can be defined as $\{v_j \in \mathcal{V} | d(v_i, v_j) \leq k\}$. However, since the traffic states are time series data and the current traffic state on a road will definitely influence the future state, we consider the all roads are self-influenced. Thus, we consider the neighborhood of a node contains the node itself and a neighborhood matrix to characterize the one-hop neighborhood relationship of the whole graph, denoted as

$$\tilde{A} = A + I \tag{1}$$

where $I$ is the identity matrix. Then, the $k$-hop neighborhood relationship of the graph nodes can be characterized by $(A + I)^k$. However, some elements in $(A + I)^k$ will inevitably exceed one. Owing to the $k$-hop neighborhood of a node is only used for describing the existence of all the $k$-hop neighbors, it is not necessary to make a node's $k$-hop neighbors weighted by the number of hops. Thus, we clip the values of all elements in $(A + I)^k$ to be in $\{0,1\}$ and define a new $k$-hop neighborhood matrix $\tilde{A}^k$, in which each element $\tilde{A}^k_{i,j}$ satisfies

$$\tilde{A}^k_{i,j} = \min\bigl((A + I)^k_{i,j}, 1\bigr) \tag{2}$$

where min refers to minimum. In this case, $\tilde{A}^1 = A^1 = A$. An intuitive example of $k$-hop neighborhood with respect to a node (a red star) is illustrated by blue points on the left side of Fig. 1.

#### 3) Free-Flow Reachable Matrix

Based on the length of each road in the traffic network, we define a distance matrix $Dist \in \mathbb{R}^{N \times N}$, where each element $Dist_{i,j}$ represents the real roadway distance from node $i$ to $j$ ($Dist_{i,i} = 0$). When taking the underlying physics of vehicle traffic on a road network into consideration, we need to understand that the impact of a roadway segment on adjacent segments is transmitted in two primary ways: 1) slowdowns and/or blockages propagating upstream; and 2) driver behavior and vehicle characteristics associated with a particular group of vehicles traveling downstream. Thus, for a traffic network-based graph or other similar graphs, the traffic impact transmission between non-adjacent nodes cannot bypass the intermediate node/nodes, and thus, we need to consider the reachability of the impact between adjacent and nearby node pairs. To ensure the traffic impact transmission between k-hop adjacent nodes follow the established traffic flow theory [53], we define a free-flow reachable matrix, $\mathcal{FFR} \in \mathbb{R}^{N \times N}$, that

$$\mathcal{FFR}_{i,j} = \begin{cases} 1, S_{i,j}^{\mathcal{FF}} m\Delta t - Dist_{i,j} \geq 0 \\ 0, \quad \text{otherwise} \end{cases}, \forall v_i, v_j \in \mathcal{V} \quad (3)$$

where $S_{i,j}^{\mathcal{FF}}$ is the free-flow speed between node $i$ and $j$, and free-flow speed [54] refers to the average speed that a motorist would travel if there were no congestion or other adverse conditions (such as severe weather). $\Delta t$ is the duration of time quantum and $m$ is a number counting how many time intervals are considered to calculate the distance travelled under free-flow speed. Thus, $m$ determines the temporal influence of formulating the $\mathcal{FFR}$. Each element $\mathcal{FFR}_{i,j}$ equals one if vehicles can traverse from node $i$ to $j$ in $m$ time-step, $m \cdot \Delta t$, with free-flow speed, and $\mathcal{FFR}_{i,j} = 0$ otherwise. Intuitively, the $\mathcal{FFR}_{i,j}$ measures whether a vehicle can travel from node $i$ to node $j$ with the free-flow speed under a specific time interval. We consider each road is self-reachable, and thus, all diagonal values of $\mathcal{FFR}$ are set as one. An example $\mathcal{FFR}$ with respect to a node (a red star) is shown by green lines on the left side of Fig. 1.

*B. Traffic Forecasting Problem*

Traffic forecasting refers to predicting future traffic states, such as traffic speed, travel time, or volume, given previously observed traffic states from a road network. In this study, the traffic network is converted into a graph consisting of all $N$ nodes, representing $N$ traffic sensing locations, and a set of edges. During a period of time $t$, the signals of these nodes representing the collected traffic states, can be denoted as $x_t \in \mathbb{R}^N$.

To formulate the traffic forecasting problem, the main aforementioned notations are summarized in the following list:

| | |
|---|---|
| $\mathcal{G}$ | Traffic network-based graph $\mathcal{G} = (\mathcal{V}, \mathcal{E})$ |
| $\mathcal{V}$ | Set of vertices in $\mathcal{G}$ with the size of $|\mathcal{V}| = N$ |
| $\mathcal{E}$ | Set of edges in $\mathcal{G}$ with the size of $|\mathcal{E}|$ |
| $A \in \mathbb{R}^{N \times N}$ | Adjacency matrix of $\mathcal{G}$ |
| $D \in \mathbb{R}^{N \times N}$ | Degree matrix of $\mathcal{G}$ |
| $\tilde{A} \in \mathbb{R}^{N \times N}$ | Neighborhood matrix defined by (1) |
| $\tilde{A}^k \in \mathbb{R}^{N \times N}$ | $k$-hop neighborhood matrix defined by (2) |
| $Dist \in \mathbb{R}^{N \times N}$ | Distance matrix |
| $\mathcal{FFR} \in \mathbb{R}^{N \times N}$ | Free-flow reachable matrix by (3) |
| $x_t \in \mathbb{R}^N$ | Vector of speed of all graph nodes at time $t$ |

The short-term traffic forecasting problem aims to learn a function $F(\cdot)$ to map $T$ time steps of historical graph signals, i.e. $\mathbf{X}_T = [x_1, \ldots, x_t, \ldots, x_T]$, to the graph signals in the subsequent one or multiple time steps. In this study, the function attempts to forecast the graph signals in the subsequent one step, i.e. $x_{T+1}$, and the formulation of $F(\cdot)$ is defined as

$$F\left([x_1, \ldots, x_t, \ldots, x_T]; \mathcal{G}(\mathcal{V}, \mathcal{E}, \tilde{A}^k, \mathcal{FFR})\right) = x_{T+1} \quad (4)$$

Further, another goal of this study is to learn the traffic impact transmission between adjacent and neighboring nodes in a traffic network-based graph by learning the weight parameters in the function $F(\cdot)$.

*C. Traffic Graph Convolution*

Previous work [24][25][28] has defined the graph convolution based the adjacency matrix. The core idea of a convolution layer in a neural network is to extract localized features from input data in a 2D or 3D matrices structure. The localized region of the input space which affects the convolution operation results is called receptive field. Analogously, the core idea of a graph convolution layer is to extract localized features from input data in a graph structure. Thus, the product of the neighborhood matrix $\tilde{A}$, the input data $x_t$, and a trainable weight matrix $W$, i.e. $\tilde{A}x_tW$, can be considered as a graph convolution operation to extract features from one-hop neighborhood [24][25]. Then, the receptive field of the graph convolution operation on a node is the one-hop neighborhood.

However, in this way, the receptive field is confined, and it only concentrates on one-hop neighboring nodes. To overcome this shortcoming, we extend the receptive field of graph convolution by replacing the one-hop neighborhood matrix $\tilde{A}$ with the $k$-hop neighborhood matrix $\tilde{A}^k$. Meanwhile, existing studies either neglect the properties of the edges in a graph, such as the distances between different sensing locations (the lengths of the graph edges) and the free-flow reachability defined in (3), or fail to consider high-order neighborhood of nodes in the graph. Hence, to comprehensively solve the network-wide forecasting problem, we consider both graph edge properties and high-order neighborhood in the traffic network-based graph. Hence, we define the $k$-order ($k$-hop) **Traffic Graph Convolution** (TGC) operation as

$$GC_t^k = \left(W_{gc_k} \odot \tilde{A}^k \odot \mathcal{FFR}\right)x_t \quad (5)$$

where $\odot$ is the Hadamard product operator, i.e. the element-wise matrix multiplication operator, and $x_t \in \mathbb{R}^N$ is the vector of traffic states (speed) of all nodes at time $t$. The $W_{gc\_k} \in \mathbb{R}^{N \times N}$ is a trainable weight matrix for the $k$-order traffic graph convolution and the $GC^k \in \mathbb{R}^N$ is the extracted $k$-order traffic graph convolution feature. Due to $\tilde{A}^k$ and $\mathcal{FFR}$ are both sparse matrices only containing 0 and 1 elements, the result of $W_{gc_k} \odot \tilde{A}^k \odot \mathcal{FFR}$ is also sparse. Further, the trained weight $W_{gc\_k}$ has the potential to measure the interactive influence between graph nodes, and thus, enhance the interpretability of the model.

In Equation (5), $k$ should be a positive integer. The larger the order $k$ is, the larger the size of the receptive field of the TGC is, and then the more neighborhood-based features can be extracted from the graph. However, $k$ is not infinite, and it can be easily proved that, for a specific graph, when increasing the value of $k$, $\tilde{A}^k \odot \mathcal{FFR}$ will eventually converge to $\mathcal{FFR}$ such that $k = K_{max}$ and $\tilde{A}^{K_{max}} \odot \mathcal{FFR} = \mathcal{FFR}$. It should be noted that, while extracting traffic graph convolution features to solve real traffic prediction problems, it is not necessary to set $k$ as the max value $K_{max}$. The trade-off between the prediction accuracy and the feature richness, which is directly related to the computational cost, should be considered and balanced.

Let $K \leq K_{max}$ denote the largest hop for traffic graph convolution in this study, and the corresponding traffic graph convolution feature is $GC_t^K$ with respect to input data $x_t$. Different hops of neighborhood in TGC will result in different extracted features. To enrich the feature space, the features extracted from different orders (from 1 to $K$) of traffic graph convolution with respect to $X_t$ are concatenated together as a vector defined as follows

$$\boldsymbol{GC}_t^{\{K\}} = [GC_t^1, GC_t^2, \ldots, GC_t^K] \quad (6)$$

The $\boldsymbol{GC}_t^{\{K\}} \in \mathbb{R}^{N \times K}$ contains all the $K$ orders of traffic graph convolutional features, as intuitively shown in the left part of Fig. 1. In this study, after operating the TGC on input data $x_t$, the generated $\boldsymbol{GC}_t^{\{K\}}$ will be fed into the following layer in the proposed neural network structure described in the following section.

### D. Comparing TGC with Spectral Graph Convolution

The proposed traffic graph convolution is based on adjacency matrix $A$, but the spectral graph convolution (SGC) is defined in the Fourier domain [50] based on the Laplacian matrix $L$, which equals

$$L = D - A \quad (7)$$

where $D$ is the degree matrix as introduced in Section III.A.2. The Laplacian matrix $L$ is symmetric positive semi-definite such that it can be diagonalized via eigen-decomposition as

$$L = U \Lambda U^T \quad (8)$$

where $\Lambda$ is a diagonal matrix containing the eigenvalues, $U$ consists of the eigenvectors, and $U^T$ is the transpose of $U$.

The spectral convolution on graph is defined as the multiplication of a signal $x_t \in \mathbb{R}^N$ with a filter $h_\theta = \text{diag}(\theta)$ parameterized by $\theta \in \mathbb{R}^N$ [24]. The $\text{diag}(\theta)$ is the diagonalized matrix given $\theta$. The spectral graph convolution operation can be described as

$$h_\theta *_G x_t = U h_\theta U^T x_t = U \text{diag}(\theta) U^T x_t \quad (9)$$

where $*_G$ is the spectral graph convolution operator. The filter $h_\theta$ that can be considered as a learnable convolutional kernel weight.

Further, for saving computational cost, the localized spectral graph convolution (LSGC) is proposed by employing a polynomial filter $h_{\theta'} = \sum_{j=0}^{k-1} \theta_j' \Lambda^j$ [23] and the learnable parameter $\theta' \in \mathbb{R}^K$. Then $K$-hop localized spectral graph convolution can be formulated as:

$$h_{\theta'} *_G x_t = U \sum_{j=0}^{K-1} \theta_j' \Lambda^j U^T x_t = \sum_{j=0}^{K-1} \theta_j' L^j x_t \quad (10)$$

The advantages of the LSGC is that it only has $K$ parameters and does not need eigen-decomposition. It is well spatial localized and each convolution operation on a centered vertex extracts the summed weighted feature of the vertex's $K$-hop neighbors. The details of SGC and LSGC can be found in the literature [22][23][24].

The comparison between TGC, SGC, and LSGC in terms of the number of parameters, computational time, and localized feature extraction, is shown in TABLE I. Comparing to SGC and LSGC, the TGC is better in terms of spatial localization because it can extract local features based on physical properties of roadways by incorporating the $\mathcal{FFR}$. TGC with more parameters has better capabilities of representing the relationships between connected nodes in the graph. Further, SGC and LSGC normally need multiple convolutional layers, which leads the SGC and LSGC to lose their interpretability. However, TGC only needs one convolution layer and its parameters can be better interpreted.

### E. Traffic Graph Convolutional LSTM

We propose a Traffic Graph Convolutional LSTM (TGC-LSTM) recurrent neural network, as shown on the right side of the Fig. 1, which learns both the complex spatial dependencies and the dynamic temporal dependencies presented in traffic data. In this model, the gates structure in the vanilla LSTM [12] and the hidden state are unchanged, but the input is replaced by the graph convolution features, which are reshaped into a vector $\boldsymbol{GC}^{\{K\}} \in \mathbb{R}^{KN}$. The forget gate $f_t$, the input gate $i_t$, the output gate $o_t$, and the input cell state $\tilde{C}_t$ in terms of time step $t$ are defined as follows

$$f_t = \sigma_g \left( W_f \cdot \boldsymbol{GC}_t^{\{K\}} + U_f \cdot h_{t-1} + b_f \right) \quad (11)$$

TABLE I
COMPARISON BETWEEN TGC, SGC, AND LSGC

| Graph convolution definition | $K$-hop TGC | SGC | $K$-hop LSGC |
|---|---|---|---|
| Graph convolution on signal $x_t$ | $(W_{gc_k} \odot \tilde{A}^K \odot \mathcal{FFR})$ | $U \text{diag}(\theta) U^T$ | $\sum_{j=0}^{K-1} \theta_j' L^j$ |
| Weight parameters | $W_{gc_k} \in \mathbb{R}^{N \times N}$ | $\theta \in \mathbb{R}^N$ | $\theta' \in \mathbb{R}^K$ |
| Computational time complexity | $O(N^2)$ | $O(N^2)$ [23] | $O(K|\mathcal{E}|)$ [23] |
| Extract Localized features | Yes. It is $k$-localized incorporating roadway physical properties. | No | Yes. It is exactly $k$-localized. |

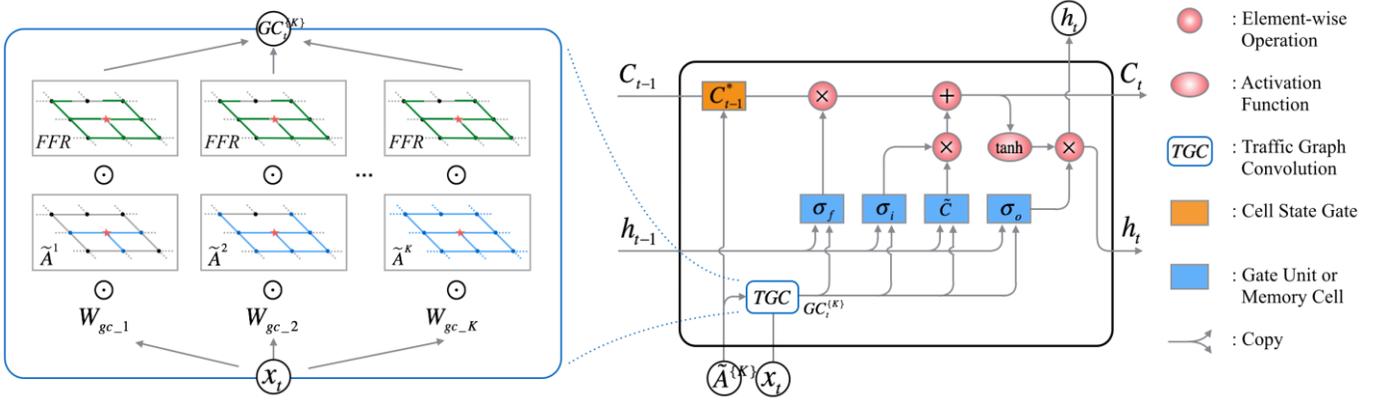

Fig. 1. The architecture of the proposed Traffic Graph Convolution LSTM is shown on the right side. The traffic graph convolution (TGC) as a component of the proposed model is shown on the left side in detail by unfolding the traffic graph convolution at time $t$, in which $\tilde{A}^k$s and $\mathcal{FFR}$ with respect to a red star node are demonstrated.

$$i_t = \sigma_g\left(W_i \cdot \boldsymbol{GC}_t^{\{K\}} + U_i \cdot h_{t-1} + b_i\right) \quad (12)$$

$$o_t = \sigma_g\left(W_o \cdot \boldsymbol{GC}_t^{\{K\}} + U_o \cdot h_{t-1} + b_o\right) \quad (13)$$

$$\tilde{C}_t = tanh\left(W_C \cdot \boldsymbol{GC}_t^{\{K\}} + U_C \cdot h_{t-1} + b_C\right) \quad (14)$$

where $\cdot$ is the matrix multiplication operator. $W_f$, $W_i$, $W_o$, and $W_C \in \mathbb{R}^{N \times KN}$ are the weight matrices, mapping the input to the three gates and the input cell state, while $U_f$, $U_i$, $U_o$, and $U_C \in \mathbb{R}^{N \times N}$ are the weight matrices for the preceding hidden state. $b_f$, $b_i$, $b_o$, and $b_C \in \mathbb{R}^N$ are four bias vectors. The $\sigma_g$ is the gate activation function, which typically is the sigmoid function, and tanh is the hyperbolic tangent function.

Due to each node in a traffic network graph is influenced by the preceding states of itself and its neighboring nodes, the LSTM cell state of each node in the graph should also be affected by neighboring cell states. Thus, a cell state gate is designed and added in the LSTM cell. The cell state gate, as shown in Fig. 1, is defined as follows

$$C_{t-1}^* = W_\mathcal{N} \odot (\tilde{A}^K \odot \mathcal{FFR}) \cdot C_{t-1} \quad (15)$$

where $W_\mathcal{N}$ is a weight matrix to measure the contributions of neighboring cell states. To correctly reflect the traffic network structure, the $W_\mathcal{N}$ is constrained by multiplying a $\mathcal{FFR}$ based $K$-hop adjacency matrix, $\tilde{A}^K \odot \mathcal{FFR}$. With this gate, the influence of neighboring cell states will be considered when the cell state is recurrently input to the subsequent time step. Then, the final cell state and the hidden state are calculated as follows

$$C_t = f_t \odot C_{t-1}^* + i_t \odot \tilde{C}_t \quad (16)$$

$$h_t = o_t \odot \tanh(C_t) \quad (17)$$

At the final time step $T$, the hidden state $h_T$ is the output of TGC-LSTM, namely the predicted value $\hat{y}_T = h_T$. Let $y_T \in \mathbb{R}^N$ denote the label of the input data $X_T \in \mathbb{R}^{N \times N}$. For the sequence prediction problem in this study, the label of time step $T$ is the input of the next time step $(T + 1)$ such that $y_T = x_{T+1}$. Then the loss during the training process is defined as

$$Loss = L(y_T, \hat{y}_T) = L(x_{T+1}, h_T) \quad (18)$$

where $L(\cdot)$ is a function to calculate the residual between the predicted value $\hat{y}_T$ and the true value $y_T$. Normally, the $L(\cdot)$ function is a Mean Squared Error (MSE) function for predicting continuous values.

To explain the proposed method in a clearer way, a pseudo-code of the TGC-LSTM algorithm is presented in Algorithm 1. Given the traffic state data $X_T$ and the graph related matrices as input, the pseudo-code mainly describes the process of generating the final output $h_T$ after $T$ steps of iteration. For simplicity, the pseudo-code does not include the mini-batch gradient descent process and the backpropagation-based parameter updating process. In Algorithm 1, Eq. is short for Equation and the function TGC-LSTM($\cdot$) refers to the whole calculation process described in Equation (11-17) in this section.

### F. Traffic Graph Convolution Regularization

Since the proposed model contains a traffic graph convolution operation, the generated set of TGC features $\boldsymbol{GC}_t^{\{K\}}$ and the learned TGC weights $\{W_{gc_1}, \dots, W_{gc_K}\}$ provide an opportunity to make the proposed model interpretable via analyzing the learned TGC weights. To confine the graph convolution features within a reasonable scale and make the learned weights more stable and interpretable, we propose two

---

**Algorithm 1** Calculation the output of the TGC-LSTM layer

**Inputs:** $X_T = [x_1, \dots, x_T]$, $\{\tilde{A}^1, \dots, \tilde{A}^K\}$, $\mathcal{FFR}$
**Parameters:** $\{W_{gc_1}, \dots, W_{gc_K}\}$, $W$s, $U$s, and $b$s in Eq. (11-14)
$\quad\quad\quad\quad W_\mathcal{N}$ in Eq. (15)
**Initialize:** $h_0 = \mathbf{0} \in \mathbb{R}^N$, $C_0 = \mathbf{0} \in \mathbb{R}^N$
$\quad$ for $t = 1$ to $T$ do
$\quad\quad$ for $k = 1$ to $K$ do
$\quad\quad\quad GC_t^k \leftarrow (W_{gc_k} \odot \tilde{A}^k \odot \mathcal{FFR}) x_t$
$\quad\quad$ end for
$\quad\quad \boldsymbol{GC}_t^{\{K\}} \leftarrow [GC_t^1, GC_t^2, \dots, GC_t^K]$
$\quad\quad h_t, C_t = \text{TGC-LSTM}(x_t, \boldsymbol{GC}_t^{\{K\}}, h_{t-1}, C_{t-1})$
$\quad$ end for
**Return:** $h_T$

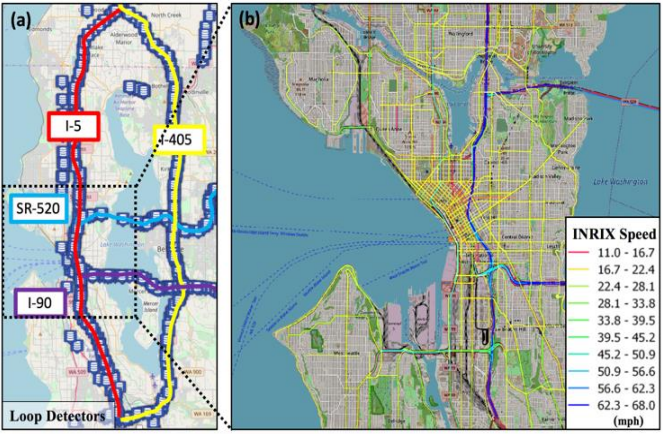

Fig. 2. (a) LOOP dataset covering the freeway network in Seattle area; (b) INRIX dataset covering the downtown Seattle area, where traffic segments are plotted with colors.

optional regularization terms that can be added to the loss function described in Equation (18).

*1) Regularization on Graph Convolution weights*

Because the graph convolution weights are not confined to be positive and each node's extracted features are influenced by multiple neighboring nodes, the graph convolution weights can vary a lot while training. Ideally, the convolution weights would be themselves informative, so that the relationships between different nodes in the network could be interpreted and visualized by plotting the convolution weights. This is not likely to be possible without regularization, because very high or low weights tend to appear somewhat randomly, with the result that high/low weights tend to cancel each other out. In combination, such weights can still represent informative features for the network, but they cannot reflect the true relationship between nodes in the graph. Thus, we add L1-norm of the graph convolution weight matrices to the loss function as a regularization term to make these weight matrices as sparse as possible. The L1 regularization term is defined as follows

$$R^{\{1\}} = \|\boldsymbol{W}_{gc}\|_1 = \sum_{i=1}^{K} |W_{gc_i}| \quad (19)$$

In this way, the trained graph convolution weight can be sparse and stable, and thus, it will be more intuitive to distinguish which neighboring node or group of nodes contribute most.

*2) Regularization on Graph Convolution features*

Considering that the impact of neighboring nodes with respect to a specific node must be transmitted through all nodes between the node of interest and the influencing node, features extracted from different hops in the graph convolution should not vary dramatically. Thus, to restrict the difference between features extracted from adjacent hops of graph convolution, an L2-norm based TGC feature regularization term is added on the loss function at each time step. The regularization term is defined as follows

$$R^{\{2\}} = \left\|\boldsymbol{GC}_T^{\{K\}}\right\|_2 = \sqrt{\sum_{i=1}^{K-1} (GC_T^i - GC_T^{i+1})^2} \quad (20)$$

In this way, the features extracted from adjacent hops of graph convolution should not differ dramatically, and thus, the graph convolution operator should be more in keeping with the physical realities of the relationships present in a traffic network.

Then, the total loss function at time $t$ can be defined as follows

$$Loss = L(h_T - x_{T+1}) + \lambda_1 R^{\{1\}} + \lambda_2 R^{\{2\}} \quad (21)$$

where $\lambda_1$ and $\lambda_2$ are penalty terms to control the weight magnitude of the regularization terms on graph convolution weights and features.

## IV. EXPERIMENTS

### A. Dataset Description

In this study, two real-world network-scale traffic speed datasets are utilized. The first contains data collected from inductive loop detectors deployed on four connected freeways (I-5, I-405, I-90, and SR-520) in the Greater Seattle Area, shown in Fig. 2 (a). This dataset, which is publicly accessible[2], contains traffic state data from 323 sensor stations over the entirety of 2015 at 5-minute intervals. The second contains road link-level traffic speeds aggregated from GPS probe data collected by commercial vehicle fleets and mobile apps provided by the company INRIX. The INRIX traffic network covers the Seattle downtown area, shown in Fig. 2 (b). This dataset describes the traffic state at 5-minute intervals for 1014 road segments and covers the entire year of 2012. We use LOOP data and INRIX data to denote these two datasets, respectively, in this study.

We adopt the speed limit as the free-flow speed, which for the segments in the LOOP traffic network is 60mph in all cases. The INRIX traffic network contains freeways, ramps, arterials, and urban corridors, and so the free-flow speeds of INRIX traffic network range from 20mph to 60mph. The distance adjacency matrices $Dist$ and free-flow reachable matrices $\mathcal{FFR}$ for both datasets are calculated based on the roadway characteristics and topology.

### B. Experimental Settings

*1) Baselines*

We compare TGC-LSTM with the following baseline models: (1) ARIMA: Auto-Regressive Integrated Moving Average model [7]; (2) SVR: Support Vector Regression [6]; (3) FNN: Feed forward neural network with two hidden layers, i.e. the multilayer perceptron, whose hidden layer size is N; (4) LSTM: Long Short-Term Memory recurrent neural network [12]; (5) DiffGRU [20]: an adjusted version of diffusion convolutional gated recurrent network [20] whose gate units are defined based on diffusion convolution. Since the graph is undirected in this study, we replace the diffusion convolution with spectral graph convolution in DiffGRU; (6) Conv+LSTM: a one-dimensional (1D) convolution layer with two channels followed by an

---

[2] https://github.com/zhiyongc/Seattle-Loop-Data

TABLE II
PERFORMANCE COMPARISON OF DIFFERENT APPROACHES. (THE NUMBER OF HOPS K IS SET AS 3 IN THE GRAPH CONVOLUTION RELATED MODEL)

| Model | LOOPf Data | | | INRIX Data | | |
|---|---|---|---|---|---|---|
| | MAE (mph)±STD | MAPE | RMSE | MAE (mph)±STD | MAPE | RMSE |
| ARIMA | 6.10± 1.09 | 13.85% | 10.65 | 4.80 ± 0.32 | 13.51% | 10.85 |
| SVR | 6.85± 1.17 | 14.39% | 11.12 | 4.78 ± 0.37 | 13.37% | 10.44 |
| FNN | 4.45± 0.81 | 10.19% | 7.83 | 2.31 ± 0.17 | 8.35% | 5.92 |
| LSTM | 2.70± 0.18 | 6.83% | 4.97 | 1.14 ± 0.09 | 3.88% | 2.43 |
| DiffGRU | 4.64±0.38 | 11.18% | 8.22 | 2.44 ± 0.09 | 8.91% | 6.34 |
| Conv+LSTM | 2.71±0.12 | 6.79% | 5.02 | 1.13 ± 0.08 | 3.80% | 2.37 |
| LSGC+LSTM | 3.16± 0.23 | 7.51% | 6.18 | 1.38 ± 0.12 | 4.54% | 2.82 |
| SGC+LSTM | 2.64± 0.12 | 6.52% | 4.80 | 1.07 ± 0.08 | 3.74% | 2.28 |
| **TGC-LSTM** | **2.57± 0.10** | **6.01%** | **4.63** | **1.02 ± 0.07** | **3.28%** | **2.18** |

LSTM layer, the 1D CNN is conducted on $x_t$ with two output channels (kernel size=5 and stride=2); (7) SGC+LSTM: stacking a one-layer spectral graph convolution layer [26] with an LSTM layer; (8) LSGC+LSTM: stacking a one-layer localized spectral graph convolution layer [23] whose $K$=3 and an LSTM layer. All the LSTM/GRU layers have the same weight dimensions. The baseline models do not include auto-encoder based models and pure CNN based models, due to the core ideas of these methodologies are totally different from the tested baseline models which are mostly single RNN layer-based models. All the neural networks are implemented based on PyTorch 1.0.1 and they are trained and evaluated on a single NVIDIA GeForce GTX 1080 Ti with 11GB memory.

*2) TGC-LSTM Model*

For both datasets, the dimensions of the hidden states of the TGC-LSTM are set as the amount of the nodes in the traffic network graphs. The size of hops in the graph convolution can vary, but we set it as 3, $K = 3$, for the model evaluation and comparison in this experiment. In this case, the $\mathcal{FFR}$ is calculated based on three time steps. The two regularization terms ($R^{\{1\}}$ and $R^{\{2\}}$) can not only confine the learnt graph convolution weights, they also can avoid overfitting causing the decrease of the prediction accuracy. Thus, there is a trade-off between the prediction accuracy and the scale of the penalty terms ($\lambda_1$ and $\lambda_2$). Based on empirically adjusting the regularization rates, the values of the $\lambda_1$ and $\lambda_2$ are both set as 0.01. We train our model by minimizing the mean square error with the batch size of 10 and the initial learning rate of $10^{-5}$. Since the RMSProp [55] can solve the gradient exploding and vanishing problems, it is used as the gradient descent optimizer whose alpha (smoothing constant) is set as 0.99 and epsilon (the term added to the denominator to improve numerical stability) is set as $10^{-8}$.

*3) Evaluation*

In this study, the samples of the input are traffic time series data with 10 time steps. The output/label is the next subsequent data of the input sequence. The performance of the proposed and the compared models are evaluated by three commonly used metrics in traffic forecasting, including 1) Mean Absolute Error (MAE), 2) Mean Absolute Percentage Error (MAPE), and 3) Root Mean Squared Error (RMSE).

$$MAE = \frac{1}{n}\sum_{i=1}^{n} |y_T - \hat{y}_T| \quad (23)$$

$$MAPE = \frac{1}{n}\sum_{i=1}^{n} \left|\frac{y_T - \hat{y}_T}{Y_T}\right| * 100\% \quad (24)$$

$$RMSE = \sqrt{\frac{1}{n}\sum_{i=1}^{n} (y_T - \hat{y}_T)^2} \quad (25)$$

*C. Experimental Results*

TABLE II demonstrates the results of the TGC-LSTM and other baseline models on the two datasets. The proposed method outperforms other models with all the three metrics on the two datasets. The ARIMA and SVR cannot compete with other methods, which suggest that non-neural-network approaches are less appropriate for this network-wide prediction task, due to the complex spatiotemporal dependencies and the high dimension features in the datasets. The basic FNN does not perform well on predicting spatial-temporal sequence. The DiffGRU performs nearly the same as the FNN. The reason might be that GRU has the no cell state to store historical information in its gate units comparing to LSTM. This can reduce the prediction capability of DiffGRU. Both LSTM and Conv+LSTM work well and they have similar performance. The SGC+LSTM performs better than vanilla LSTM, which demonstrates the feature extraction by using spectral graph convolution is beneficial for traffic forecasting. However, the LSGC+LSTM does not outperform LSTM resulting from utilizing one-layer LSGC, whose parameters is not enough for representing the network features. The proposed TGC-LSTM, which capture graph-based features while accommodating the physical specialties of traffic networks, performs better than all other approaches. It should be noted that, for the INRIX data, during the nighttime or off-peak hours when there are no observed speed values on specific roads, the missing speed values are comprehensively imputed by the data provider. Thus, there are few variations at the non-peak hours in the INRIX data. Further, the speed values in the INRIX data are all integers. Therefore, the calculated errors of the INRIX data is less than that of the LOOP data and the evaluated performance on INRIX data is inflated somewhat.

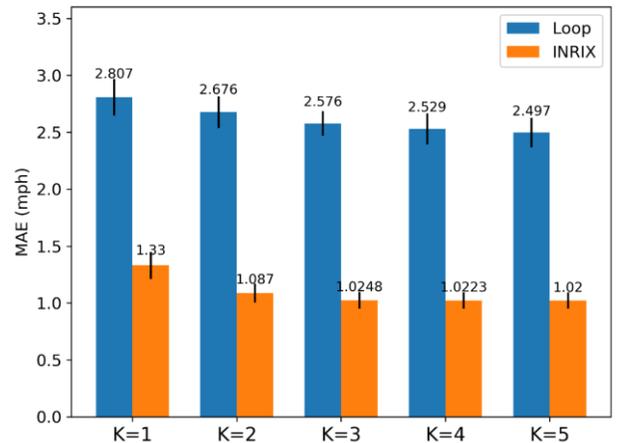

Fig. 3. Histogram of performance comparison for the influence of orders (hops) of graph convolution in the TGC-LSTM on INRIX and LOOP datasets.

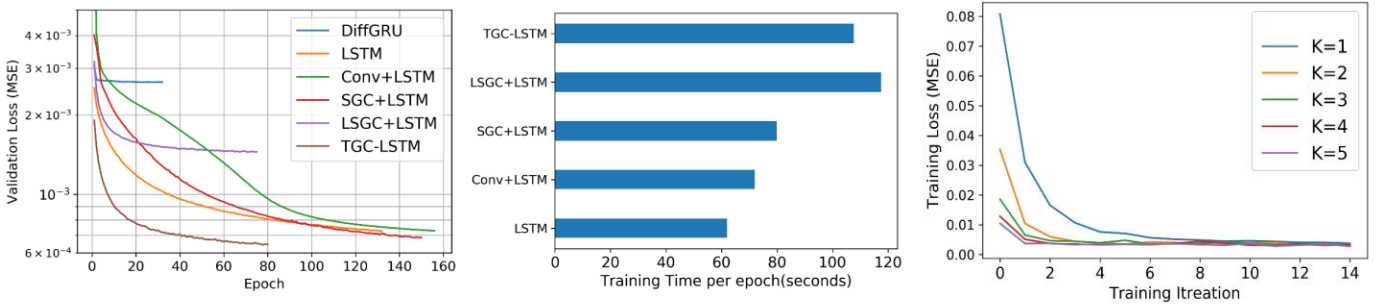

Fig. 4. (a) Validation loss versus training epoch (batch size = 40 and early stopping patience = 10 epochs). (b) Histogram of model's training time per epochs. (c) Compare training efficiency with different $K$ hops of TGC: training loss versus training iteration (batch size = 40). (The figures are generated based on the LOOP data.)

Fig. 3 shows a histogram of performance comparison on the effects of orders (hops) of the graph convolution in the TGC-LSTM. The model performance is improved when the value of $K$ increases. For the LOOP data, the performance improves slightly when $K$ is gradually increased. But for the INRIX data, there is a big improvement in when $K$ increases to two from one. The complex structure and the various road types in the INRIX traffic network could be the main reason for this performing difference. Further, when $K$ is larger than two, the improvement of the prediction is quite limited. This is also the reason why we choose $K=3$ in the model comparison part, as shown in TABLE II.

### D. Training Efficiency

In this subsection, we compare the training efficiency of the proposed model and other LSTM-based models. Fig. 4 (a) shows the validation loss curves versus the training epoch. Due to the early stopping mechanism is used in the training process, the numbers of training epochs are different. The TGC-LSTM needs less epochs to converge than the SGC+LSTM and the LSGC+LSTM. In addition, the loss of the TGC-LSTM decreases fastest among the compared models. Fig. 4 (b) shows the comparison of the training time per epoch of different models. The training cost of Conv+LSTM is between that of LSTM and SGC+LSTM. TGC-LSTM costs twice as much as LSTM does. The time required for SGC+LSTM is less than that for TGC-LSTM, while LSGC+LSTM costs slightly more than TGC-LSTM. Fig. 4 (c) shows the training losses of TGC-LSTM with different hops of graph convolution components. The rate of convergence increases when increasing the number of hops, $k$. In our experiments, when $k$ is larger than 3, the training and validation results improve only marginally for both INRIX and LOOP datasets.

### E. Effect of Regularization

The model's loss function can add regularization terms to avoid overfitting. The proposed L1-norm on the graph convolution weights and L2-norm on the graph convolutional features can further help the model to confine the learned weights and features. However, there is a trade-off between the prediction accuracy and the scale of the penalty terms ($\lambda_1$ and $\lambda_2$). As tested, by adding the regularization terms to the loss function with the penalty rates setting as 0.01, the MAEs of the proposed model tested on the two datasets increase around 0.02, which are still superior to baseline models. Meanwhile, the TGC weight sparsity is increased and the value of the feature regularization $R^{\{2\}}$ is lower than that of the proposed model without regularization terms in the loss function, which means the TGC features' consistency is enhanced. Thus, it is worth adding these regularization terms to the loss function to help the trained model to be more interpretable. Fig. 5 (a) and (b) show portions of the averaged graph convolution weight matrices for the INRIX data and the LOOP data, respectively, where $K = 3$ and the average weight is calculated by $\frac{1}{K}\sum_{i=1}^{K} W_i \odot \tilde{A}^i \odot \mathcal{FFR}$. The road segment names, which are not displayed, are aligned on the vertical and horizontal axes with the same order in each figure. The colored dots in the matrices in Fig. 5 (a) and (b) illustrate the weight of the contribution of a single node to its neighboring nodes. Since we align the traffic states of roadway segments based on their interconnectedness in the training data, most of the weights are distributed around the diagonal line of the weight matrix. The INRIX network is more complex and the average degree of nodes in the INRIX graph is higher than that in the LOOP graph. Hence, the dots in the average weight matrix of the INRIX graph convolution are more scattered. But these dots still form multiple clusters demonstrating the weights of several nearby or connected road segments. Considering roadway segments are influenced by their neighboring or nearby connected segments, the nodes with the large absolute weight in a cluster are very likely to be key road segments in the local traffic network. In this way, we can infer the bottlenecks of the traffic network from the traffic graph convolution weight matrices.

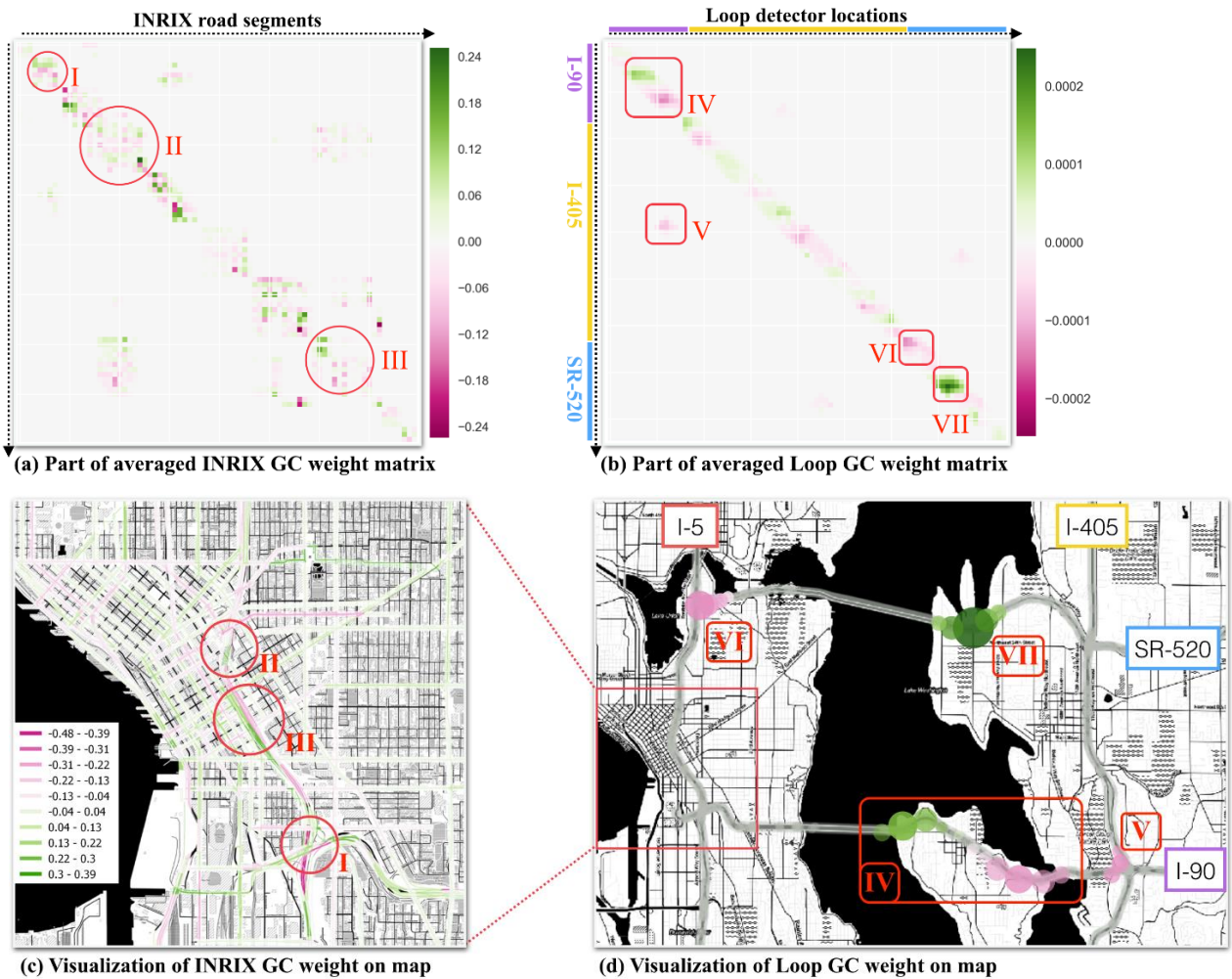

Fig. 5. (a) Visualization of a proportion of the INRIX GC weight matrix, in which three representative weight areas are tagged. (b) Visualization of a proportion of the LOOP GC weight matrix, in which four representative weight areas are tagged. (c) Visualization of the INRIX graph convolution weight on the real traffic network using colored lines. (d) Visualization of the four tagged weight areas in the LOOP graph convolution weight on the Seattle freeway network using colorful circles.

### F. Model Interpretation and Visualization

To better understand the contribution of the graph convolution weight, we mark seven groups of representative weights in Fig. 5 (a) and (b) and visualize their physical locations on the real map in Fig. 5 (c) and (d), by highlighting them with Roman numerals and red boxes. The influence of these marked weights on neighboring nodes in the INRIX and LOOP data are visualized by lines and circles, respectively, considering the INRIX traffic network is too dense to use circles. The darkness of the green and pink colors and the sizes of the circles represent the magnitude of influence. It should be noted that the darkness of colors on lines on the INRIX map and the size of the circles on the LOOP map will change when the model is trained with different scales of regularization terms ($\lambda_1$ and $\lambda_2$).

From Fig. 5 (c), we can find the marked areas with dark colors in the INRIX GC weight matrix, (I), (II), and (III), are all located at very busy and congested freeway entrance and exit ramps in Seattle downtown area. In Fig. 5 (d), the area tagged with (IV) is quite representative because the two groups of circles are located at the intersections between freeways and two main corridors that represent the entrances to an island (Mercer Island). Areas (V) and (VI) are the intersections between I-90 and I-405 and between I-5 and SR-520, respectively. The VII area located on SR-520 contains a frequent-congested ramp connecting to the city of Bellevue, the location of which is highlighted by the biggest green circle. Additionally, there are many other representative areas in the graph convolution weight matrix, but we cannot show all of them due to space limits. By comparing the weight matrix with

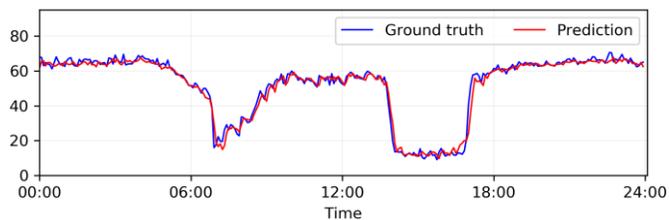
(a) Sensor ID: d005es16756 in LOOP dataset on 2015-01-04.

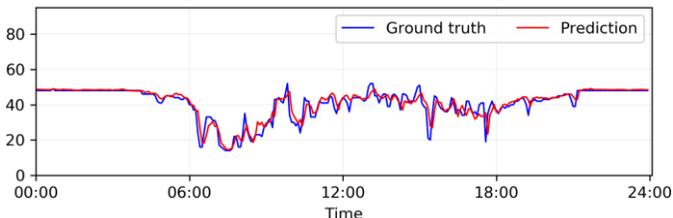
(b) Sensor ID: 114P13774 in INRIX dataset on 2012-01-03.

Fig. 6. Traffic time series forecasting visualization for LOOP and INRIX datasets on two randomly selected days.

the physical realities of the traffic network, it can be shown that the proposed method effectively captures spatial dependencies and helps to identify the most influential points/segments in the traffic network.

Fig. 6 visualizes the predicted traffic speed sequences and the ground truth of two locations selected from the LOOP and INRIX dataset. Though the traffic networks of the two datasets are very different, the curves demonstrate that the trends of the traffic speed are predicted well at both peak traffic and off-peak hours.

## V. Conclusion

In this paper, we learn the traffic network as a graph and define a traffic graph convolution operation to capture spatial features from the traffic network. The traffic graph convolution incorporates the adjacency matrix and the proposed free-flow reachable matrix to extract localized features from the graph. We propose a traffic graph convolutional LSTM neural network to forecast network-wide traffic states. We also design two regularization terms on the TGC weights and TGC features, respectively, that can be added to the model's loss function to help the learned TGC weight to be more stable and interpretable. By evaluating on two real-world traffic datasets, our approach is proved to be superior to the compared baseline models. In addition, the learned TGC weight can help to identify the most influential roadways, and thus, enhance the interpretability of the proposed model.

For future work, we will move forward to improve the model's prediction performance in terms of accuracy and robustness, and further investigate how to conduct the convolution on both spatial and temporal dimensions to make the neural network more interpretable.

## References

[1] B. Yu, H. Yin, and Z. Zhu, "Spatio-temporal graph convolutional neural network: A deep learning framework for traffic forecasting," *arXiv Prepr. arXiv1709.04875*, 2017.

[2] D. Park and L. R. Rilett, "Forecasting freeway link travel times with a multilayer feedforward neural network," *Comput. Civ. Infrastruct. Eng.*, vol. 14, no. 5, pp. 357–367, 1999.

[3] E. I. Vlahogianni, M. G. Karlaftis, and J. C. Golias, "Short-term traffic forecasting: Where we are and where we're going," *Transp. Res. Part C Emerg. Technol.*, vol. 43, pp. 3–19, Jun. 2014.

[4] X. Ma, Z. Tao, Y. Wang, H. Yu, and Y. Wang, "Long short-term memory neural network for traffic speed prediction using remote microwave sensor data," *Transp. Res. Part C Emerg. Technol.*, vol. 54, pp. 187–197, May 2015.

[5] X. Ma, Z. Dai, Z. He, J. Ma, Y. Wang, and Y. Wang, "Learning traffic as images: a deep convolutional neural network for large-scale transportation network speed prediction," *Sensors*, vol. 17, no. 4, p. 818, 2017.

[6] A. J. Smola and B. Schölkopf, "A tutorial on support vector regression," *Stat. Comput.*, vol. 14, no. 3, pp. 199–222, 2004.

[7] M. M. Hamed, H. R. Al-Masaeid, and Z. M. B. Said, "Short-term prediction of traffic volume in urban arterials," *J. Transp. Eng.*, vol. 121, no. 3, pp. 249–254, 1995.

[8] X. Ma, Z. Tao, Y. Wang, H. Yu, and Y. Wang, "Long short-term memory neural network for traffic speed prediction using remote microwave sensor data," *Transp. Res. Part C Emerg. Technol.*, vol. 54, pp. 187–197, 2015.

[9] J. Van Lint, S. Hoogendoorn, and H. Van Zuylen, "Freeway travel time prediction with state-space neural networks: modeling state-space dynamics with recurrent neural networks," *Transp. Res. Rec. J. Transp. Res. Board*, no. 1811, pp. 30–39, 2002.

[10] W. Huang, G. Song, H. Hong, and K. Xie, "Deep Architecture for Traffic Flow Prediction: Deep Belief Networks With Multitask Learning.," *IEEE Trans. Intell. Transp. Syst.*, vol. 15, no. 5, pp. 2191–2201, 2014.

[11] Y. Lv, Y. Duan, W. Kang, Z. Li, F.-Y. Wang, and others, "Traffic flow prediction with big data: A deep learning approach.," *IEEE Trans. Intell. Transp. Syst.*, vol. 16, no. 2, pp. 865–873, 2015.

[12] S. Hochreiter and J. Schmidhuber, "Long Short-Term Memory," *Neural Comput.*, vol. 9, no. 8, pp. 1735–1780, Nov. 1997.

[13] K. Cho *et al.*, "Learning phrase representations using RNN encoder-decoder for statistical machine translation," *arXiv Prepr. arXiv1406.1078*, 2014.

[14] Z. Cui, R. Ke, and Y. Wang, "Deep Stacked Bidirectional and Unidirectional LSTM Recurrent Neural Network for Network-wide Traffic Speed Prediction," in *6th International Workshop on Urban Computing (UrbComp 2017)*, 2016.

[15] R. Yu, Y. Li, C. Shahabi, U. Demiryurek, and Y. Liu, "Deep learning: A generic approach for extreme condition traffic forecasting," in *Proceedings of the 2017 SIAM International Conference on Data Mining*,

2017, pp. 777–785.
[16] F. Kong, J. Li, B. Jiang, T. Zhang, and H. Song, "Big data-driven machine learning-enabled traffic flow prediction," *Trans. Emerg. Telecommun. Technol.*, p. e3482.
[17] J. Zhang, Y. Zheng, and D. Qi, "Deep Spatio-Temporal Residual Networks for Citywide Crowd Flows Prediction.," in *AAAI*, 2017, pp. 1655–1661.
[18] H. Yu, Z. Wu, S. Wang, Y. Wang, and X. Ma, "Spatiotemporal recurrent convolutional networks for traffic prediction in transportation networks," *Sensors*, vol. 17, no. 7, p. 1501, 2017.
[19] B. Yu, M. Li, J. Zhang, and Z. Zhu, "3D Graph Convolutional Networks with Temporal Graphs: A Spatial Information Free Framework For Traffic Forecasting," Mar. 2019.
[20] Y. Li, R. Yu, C. Shahabi, and Y. Liu, "Diffusion Convolutional Recurrent Neural Network: Data-Driven Traffic Forecasting," in *International Conference on Learning Representations (ICLR '18)*, 2018.
[21] J. Atwood and D. Towsley, "Diffusion-convolutional neural networks," in *Advances in Neural Information Processing Systems*, 2016, pp. 1993–2001.
[22] J. Bruna, W. Zaremba, A. Szlam, and Y. LeCun, "Spectral Networks and Locally Connected Networks on Graphs," *arXiv Prepr. arXiv1312.6203*, Dec. 2013.
[23] M. Defferrard, X. Bresson, and P. Vandergheynst, "Convolutional Neural Networks on Graphs with Fast Localized Spectral Filtering," in *Advances in Neural Information Processing Systems*, 2016, pp. 3844–3852.
[24] T. N. Kipf and M. Welling, "Semi-supervised classification with graph convolutional networks," in *International Conference on Learning Representations (ICLR '17)*, 2017.
[25] Z. Zhou and X. Li, "Convolution on Graph: A High-Order and Adaptive Approach," *arXiv Prepr. arXiv1706.09916*, 2018.
[26] M. Henaff, J. Bruna, and Y. LeCun, "Deep Convolutional Networks on Graph-Structured Data," *arXiv Prepr. arXiv1506.05163*, Jun. 2015.
[27] M. Simonovsky and N. Komodakis, "Dynamic edgeconditioned filters in convolutional neural networks on graphs," in *Proc. CVPR*, 2017.
[28] S. Abu-El-Haija *et al.*, "MixHop: Higher-Order Graph Convolution Architectures via Sparsified Neighborhood Mixing," *Int. Conf. Mach. Learn.*, 2019.
[29] N. G. Polson and V. O. Sokolov, "Deep learning for short-term traffic flow prediction," *Transp. Res. Part C Emerg. Technol.*, vol. 79, pp. 1–17, 2017.
[30] J. Hua and A. Faghri, "Applcations of Artificial Neural Networks to Intelligent Vehicle-Highway Systems," *Record (Washington).*, vol. 1453, p. 83, 1994.
[31] H. Yin, S. Wong, J. Xu, and C. K. Wong, "Urban traffic flow prediction using a fuzzy-neural approach," *Transp. Res. Part C Emerg. Technol.*, vol. 10, no. 2, pp. 85–98, 2002.
[32] F. Kong, J. Li, B. Jiang, and H. Song, "Short-term traffic flow prediction in smart multimedia system for Internet of Vehicles based on deep belief network," *Futur. Gener. Comput. Syst.*, vol. 93, pp. 460–472, 2019.
[33] B. Liao *et al.*, "Deep Sequence Learning with Auxiliary Information for Traffic Prediction," in *Proceedings of the 24th ACM SIGKDD International Conference on Knowledge Discovery & Data Mining*, 2018, pp. 537–546.
[34] Y. Liang, Z. Cui, Y. Tian, H. Chen, and Y. Wang, "A Deep Generative Adversarial Architecture for Network-Wide Spatial-Temporal Traffic-State Estimation," *Transp. Res. Rec. J. Transp. Res. Board*, p. 036119811879873, Oct. 2018.
[35] Y. Lin, X. Dai, L. Li, and F.-Y. Wang, "Pattern Sensitive Prediction of Traffic Flow Based on Generative Adversarial Framework," *IEEE Trans. Intell. Transp. Syst.*, no. 99, pp. 1–6, 2018.
[36] Y. Duan, Y. Lv, and F.-Y. Wang, "Travel time prediction with LSTM neural network," in *Intelligent Transportation Systems (ITSC), 2016 IEEE 19th International Conference on*, 2016, pp. 1053–1058.
[37] Z. Zhao, W. Chen, X. Wu, P. C. Y. Chen, and J. Liu, "LSTM network: a deep learning approach for short-term traffic forecast," *IET Intell. Transp. Syst.*, vol. 11, no. 2, pp. 68–75, 2017.
[38] X. Song, H. Kanasugi, and R. Shibasaki, "DeepTransport: Prediction and Simulation of Human Mobility and Transportation Mode at a Citywide Level.," in *IJCAI*, 2016, pp. 2618–2624.
[39] X. Ma, Y. Li, Z. Cui, and Y. Wang, "Forecasting Transportation Network Speed Using Deep Capsule Networks with Nested LSTM Models," *arXiv Prepr. arXiv1811.04745*, 2018.
[40] Y. Wu, H. Tan, L. Qin, B. Ran, and Z. Jiang, "A hybrid deep learning based traffic flow prediction method and its understanding," *Transp. Res. Part C Emerg. Technol.*, vol. 90, pp. 166–180, May 2018.
[41] R. Ke, W. Li, Z. Cui, and Y. Wang, "Two-Stream Multi-Channel Convolutional Neural Network (TM-CNN) for Multi-Lane Traffic Speed Prediction Considering Traffic Volume Impact," *arXiv Prepr. arXiv1903.01678*, 2019.
[42] C. Zhang and P. Patras, "Long-term mobile traffic forecasting using deep spatio-temporal neural networks," in *Proceedings of the Eighteenth ACM International Symposium on Mobile Ad Hoc Networking and Computing*, 2018, pp. 231–240.
[43] D. Zhang and M. R. Kabuka, "Combining weather condition data to predict traffic flow: a GRU-based deep learning approach," *IET Intell. Transp. Syst.*, vol. 12, no. 7, pp. 578–585, 2018.
[44] W. Jin, Y. Lin, Z. Wu, and H. Wan, "Spatio-Temporal Recurrent Convolutional Networks for Citywide Short-term Crowd Flows Prediction," in *Proceedings of the 2nd International Conference on Compute and Data Analysis*, 2018, pp. 28–35.
[45] C. Chen *et al.*, "Exploiting Spatio-Temporal Correlations with Multiple 3D Convolutional Neural

Networks for Citywide Vehicle Flow Prediction," in *2018 IEEE International Conference on Data Mining (ICDM)*, 2018, pp. 893–898.
[46] S. Guo, Y. Lin, S. Li, Z. Chen, and H. Wan, "Deep Spatial-Temporal 3D Convolutional Neural Networks for Traffic Data Forecasting," *IEEE Trans. Intell. Transp. Syst.*, 2019.
[47] Y. Sun, X. Yu, R. Bie, and H. Song, "Discovering time-dependent shortest path on traffic graph for drivers towards green driving," *J. Netw. Comput. Appl.*, vol. 83, pp. 204–212, 2017.
[48] H. Sun, J. Wu, D. Ma, and J. Long, "Spatial distribution complexities of traffic congestion and bottlenecks in different network topologies," *Appl. Math. Model.*, vol. 38, no. 2, pp. 496–505, 2014.
[49] G. Kalafatas and S. Peeta, "An exact graph structure for dynamic traffic assignment: Formulation, properties, and computational experience," 2007.
[50] D. I. Shuman, S. K. Narang, P. Frossard, A. Ortega, and P. Vandergheynst, "The emerging field of signal processing on graphs: Extending high-dimensional data analysis to networks and other irregular domains," *IEEE Signal Process. Mag.*, vol. 30, no. 3, pp. 83–98, 2013.
[51] X. Geng *et al.*, "Spatiotemporal multi-graph convolution network for ride-hailing demand forecasting," in *2019 AAAI Conference on Artificial Intelligence (AAAI'19)*, 2019.
[52] Q. Zhang, Q. Jin, J. Chang, S. Xiang, and C. Pan, "Kernel-Weighted Graph Convolutional Network: A Deep Learning Approach for Traffic Forecasting," in *2018 24th International Conference on Pattern Recognition (ICPR)*, 2018, pp. 1018–1023.
[53] C. F. Daganzo, "The cell transmission model: A dynamic representation of highway traffic consistent with the hydrodynamic theory," *Transp. Res. Part B Methodol.*, vol. 28, no. 4, pp. 269–287, 1994.
[54] K. Hunter-Zaworski *et al.*, "Transportation engineering online lab manual." 2003.
[55] T. Tieleman and G. Hinton, "Lecture 6.5-rmsprop: Divide the gradient by a running average of its recent magnitude," *COURSERA Neural networks Mach. Learn.*, vol. 4, no. 2, pp. 26–31, 2012.